\def\mytitle{Walking Stabilization Using Step Timing and Location Adjustment on the Humanoid Robot, Atlas}
\def\myauthor{Robert J. Griffin\textsuperscript{1,2}, Georg Wiedebach\textsuperscript{2}, Sylvain Bertrand\textsuperscript{2}, Alexander Leonessa\textsuperscript{1}, Jerry Pratt\textsuperscript{2}
\thanks{This work was funded through the NSF NRI Grant No. 1525972 and by the National Aeronautics and Space Administration Grant No. NNX12AP97G.}
\thanks{\textsuperscript{1} The author is with the Terrestrial Robotics Engineering \& Controls Lab, Virginia Tech, 635 Prices Fork Rd, Blacksburg, VA 24060, United States }
\thanks{\textsuperscript{2} The author is with the Florida Institute for Human and Machine Cognition, 40 S Alcaniz St, Pensacola, FL 32502, United States}
\thanks{Email : \url{ {rgriffin, sbertrand, gwiedebach, jpratt}@ihmc.us}, \url{{leonessa@vt.edu}}
}}
\def\mycolumns{twocolumn}
\def\mydocclass{conference}
\def\myabstract{abstract}
\def\bibliocommand{\bibliography{mybib}}
\def\mypackages{packages}

\ifx\mydocclass\undefined
\def\mydocclass{conference}
\fi

\ifx\mycolumns\undefined
\def\mycolumns{onecolumn}
\fi

\ifx\mysize\undefined
\def\mysize{10pt}
\fi

\ifdefined\myinnermargin

\fi
\ifdefined\myoutermargin

\fi
\ifdefined\mytopmargin

\fi
\ifdefined\mybottommargin

\fi

\documentclass[\mysize,\mydocclass,\mycolumns]{ieeeconf}
\IEEEoverridecommandlockouts
\usepackage{graphics}
\usepackage{graphicx}
\usepackage{url}
\usepackage{hyperref}
\usepackage{apalike}
\usepackage{tabulary}
\usepackage{booktabs}

\ifx\bibliocommand\undefined
\else

\usepackage[
    backend=biber,
    natbib=true,
    style=ieee,
    url=false,
    doi=false,
    isbn=false,
    safeinputenc=true
]{biblatex}
\AtEveryBibitem{\clearfield{issn}}
\renewcommand{\bibliography}[1]{\addbibresource{#1.bib}}
\bibliocommand

\fi

\ifdefined\mypackages
\usepackage{array}
\usepackage{amsmath}
\usepackage{amsfonts}
\usepackage{algorithm}
\usepackage{algpseudocode}

\usepackage{dirtytalk}
\usepackage{xcolor}
\usepackage[english]{babel}

\newcommand{\V}[1] {\boldsymbol{\mathbf{#1}}}

\def\IEEEeqnarraybox{\@IEEEeqnarraystarformfalse\ifmmode\@IEEEeqnarrayboxHBOXSWfalse\else\@IEEEeqnarrayboxHBOXSWTRUE\fi%
\@IEEEeqnarraybox}
\def\endIEEEeqnarraybox{\end@IEEEeqnarraybox}

\newcommand{\arraybegin}[1]{\begin{IEEEeqnarraybox*}[][c]{#1}}
\newcommand{\arrayend}{\end{IEEEeqnarraybox*}}

\fi

\renewcommand\[{\begin{equation}}
\renewcommand\]{\end{equation}}

\title{\mytitle}

\ifx\myassociation\undefined
\author{\myauthor}
\else
\author{\myauthor \thanks{\myassociation}}
\fi

\begin{document}

\maketitle
\ifx\myabstract\undefined
\else
\begin{abstract}
While humans are highly capable of recovering from external disturbances and uncertainties that result in large tracking errors, humanoid robots have yet to reliably mimic this level of robustness.
Essential to this is the ability to combine traditional \say{ankle strategy} balancing with step timing and location adjustment techniques.
In doing so, the robot is able to step quickly to the necessary location to continue walking.
In this work, we present both a new swing speed up algorithm to adjust the step timing, allowing the robot to set the foot down more quickly to recover from errors in the direction of the current capture point dynamics, and a new algorithm to adjust the desired footstep, expanding the base of support to utilize the center of pressure (CoP)-based ankle strategy for balance.
We then utilize the desired centroidal moment pivot (CMP) to calculate the momentum rate of change for our inverse-dynamics based whole-body controller.
We present simulation and experimental results using this work, and discuss performance limitations and potential improvements.
\end{abstract}
\fi

\ifx\mykeywords\undefined
\else
\fi

\section{Introduction}
\label{introduction}

People are very adept at recovering from large disturbances and uncertainties when walking.
Shifting the Center of Pressure (CoP) within the available foothold (the \say{ankle strategy}) is common, as is using angular momentum, by lunging the upper body (the \say{hip strategy})~\citep{Horak_1986} or windmilling the arms~\citep{Pijnappels_2010}.
Angular momentum has its limits, though, and the control authority of the ankle strategy decreases as the walking speed increases and becomes more dynamic.
To handle these limitations, humans quickly adjust their step to the right location and continue walking~\citep{Maki_1997}.

Humanoid robots can, in theory, utilize these same approaches, but have yet to match the speed and adaptability of humans.
Robots have been demonstrated to be very capable of walking using a set of desired footsteps, stably tracking desired center of mass (CoM) motions, as long as the tracking error does not become too large.
This has primarily been performed by controlling either the Zero Moment Point (ZMP), Instantaneous Capture Point (ICP), or Divergent Component of Motion (DCM) with momentum based methods.
The Linear Inverted Pendulum Model (LIPM) has been well utilized to generate feasible CoM motions using analytic solutions~\citep{Kajita_2001}, preview control~\citep{Kajita_2003}, and Differential Dynamic Program~\citep{Feng_2014}, among others.
Both the ICP~\citep{Pratt_2006} and DCM~\citep{Takenaka_2009} were introduced by splitting the LIPM dynamics into stable and unstable components, and then controlling only this unstable portion to maintain balance.
The LIPM dynamics have then been tracked successfully using momentum-based whole-body control techniques with both traditional feedback controllers~\citep{Feng_2014} and LQR-based methods~\citep{Kuindersma_2014}.
ICP and DCM methods have also been used with whole-body controllers to effectively stabilize the walking motion~\citep{Pratt_2012, Koolen_2016b, Hopkins_2015}.
Due to the limited size of the support polygon, however, these type of tracking controllers are ill equipped to handle large tracking errors, and have very limited effective control authority.
While angular momentum has been illustrated as providing additional controllability~\citep{Wiedebach_2016}, further improvements are still needed to handle the large tracking errors that may result from external disturbances and uncertainties.

\begin{figure}[!t]
\centering
\includegraphics[width=3.0in]{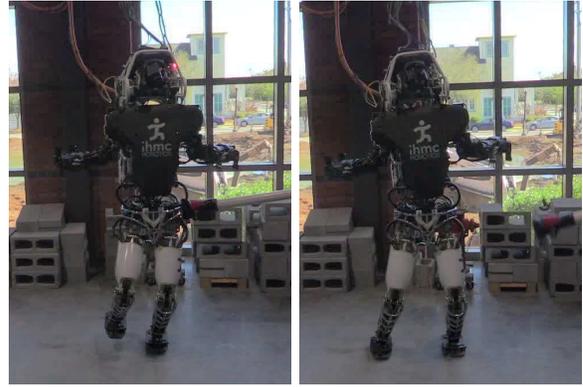}
\vspace{-2mm}
\caption{
Atlas recovering from a lateral push while stepping in place.
}
\vspace{-8mm}
\label{fig:side_push_screen_capture}
\end{figure}

To improve robustness in the face of large errors, several authors have mimicked nature and introduced step adjustment algorithms.
Some works have formulated model predictive controllers (MPC) as quadratic programs to achieve this step adjustment~\citep{Diedam_2008, Herdt_2010, Griffin_2015}.
In~\citep{Diedam_2008, Herdt_2010}, the step locations are optimized to reject disturbances using the ZMP dynamics while minimizing the CoM jerk to ensure smooth motions.
Instead of utilizing the ZMP dynamics, the MPC in~\citep{Griffin_2015} is based on the DCM dynamics, but similarly optimizes footstep locations while trying to provide \say{nice} CoM motions.
Alternatively,~\citep{Feng_2016} simply uses the LIPM dynamics to determine the necessary upcoming footstep to return to the desired step plan.
This is similar to the work in~\citep{Englsberger_2015}, which integrates the current DCM forward in time to calculate the necessary footstep location to return to the nominal trajectory.
While highly efficient, as they are not optimizing full trajectories, neither~\citep{Feng_2016} or ~\citep{Englsberger_2015} consider the combined effects of the ankle strategy with step adjustment.

Instead of adjusting the footstep location, however, the foot can also simply be set down more quickly, another common action employed by humans.
However, adjusting the step timing is a challenge, as it tends to result in nonlinearities, and so it is typically viewed as fixed.
~\citep{Kryczka_2015} uses a nonlinear optimization-based pattern generator to find the optimized step positions and step timing given the current CoM state, which are then tracked using a ZMP based feedback controller.
~\citep{Aftab_2012} augments the earlier work of~\citep{Herdt_2010} by allowing the step time to vary, as well, but again utilizes nonlinear optimization to do so.
Instead, ~\citep{Khadiv_2016} approximates this nonlinear term as a linear one, allowing the problem to maintain its convexity and efficiency.
These methods all, however, use optimization to determine the timing adjustment.
We believe that the advantages of timing adjustment can be captured using only the ICP dynamics.

In this work, we present a simple timing adjustment algorithm that is highly effective when the ICP tracking error is in the direction of the desired motion, essentially speeding up the dynamic plan in the direction of this error.
This then greatly improves the effectiveness of the disturbance rejection with step adjustment, as the robot is able to quickly step to the necessary location for recovery.
For step adjustment, instead of using traditional MPC techniques that optimize the entire trajectory, we instead combine the ability to utilize CoP control like in~\citep{Herdt_2010} with step adjustment to return to the nominal ICP plan, as in~\citep{Englsberger_2015}.
This can be done by observing that the reference ICP trajectory is a linear function of the upcoming footstep locations.
Then, by embedding a proportional feedback controller into a quadratic program, the reference trajectory can be optimized by adjusting the footsteps, taking into account the CoP feedback control action.
This makes for a highly efficient algorithm that can be run on robotic hardware in real-time at high frequencies.

\section{Dynamic Planning and Control}
\label{dynamicplanningandcontrol}

The underlying dynamic planning algorithm utilized on Atlas is based on the ICP, and is fully described in~\citep{Englsberger_2014}.
Note that in~\citep{Englsberger_2014}, the authors utilize the DCM, but, assuming constant height, this is formulaically equivalent in $x$-$y$ to the ICP.
We will summarize this approach in the following paragraphs.

The ICP is a transformation of the CoM state defined as
\[
\V{\xi} = \V{x} + \frac{ 1 }{ \omega_0 } \dot{\V{x}},
\label{eqn:icp_definition}
\]
where $\V{\xi} = \left[ \xi_x, \ \xi_y \right]^T$ is the ICP position, $\V{x} = \left[ x, \ y \right]^T$ and $\dot{\V{x}} = \left[ \dot{x}, \ \dot{y} \right]^T$ are the CoM position and velocity, and $\omega_0 = \sqrt{g / \Delta z_{\text{com}}}$ is the natural frequency of the inverted pendulum. By reordering this, we can see that the CoM has stable first order dynamics with respect to the ICP, meaning that it will converge to the ICP over time.
Through differentiation, the ICP dynamics are defined as
\[
\dot{\V{\xi}} = \omega_0 \left( \V{\xi} - \V{r}_{\text{cmp}} \right),
\label{eqn:icp_dynamics}
\]
where we see that the Centroidal Moment Pivot (CMP) point~\citep{Popovic_2005}, $\V{r}_{\text{cmp}}$, controls the ICP dynamics.
From \autoref{eqn:icp_dynamics}, the CMP is defined as
\[
\V{r}_{\text{cmp}} = \V{\xi} - \frac{1}{\omega_0} \dot{\V{\xi}},
\label{eqn:cmp_definition}
\]
allowing it to be calculated from a given ICP trajectory.

\subsection{Dynamic Planning}
\label{dynamicplanning}

\begin{figure}[!t]
\centering
\includegraphics[width=3.2in]{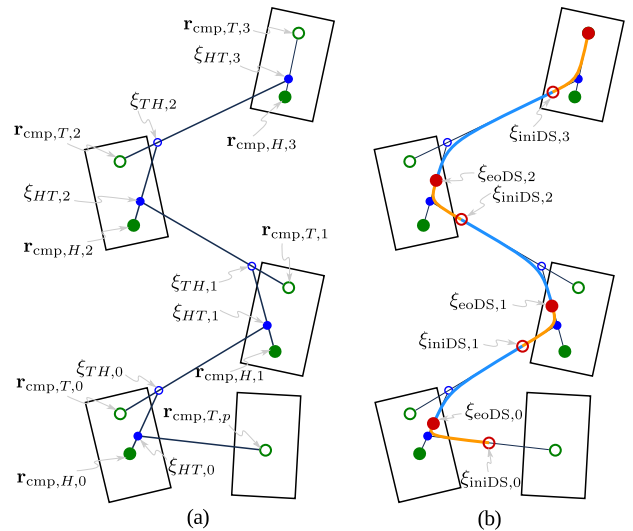}
\vspace{-3mm}
\caption{Heel-to-Toe ICP trajectory~\citep{Englsberger_2014}, with left representing instantaneous shifting between heel and toe and CMPs, and right using smoothing splines.}
\vspace{-6mm}
\label{fig:recursive_icp_plan}
\end{figure}

From the definition of the ICP dynamics in \autoref{eqn:icp_dynamics}, the linear, first order differential equation has a closed form solution
\[
\V{\xi}(t) = e^{\omega_0 t} \left( \V{\xi}_0 - \V{r}_{\text{cmp}} \right) + \V{r}_{\text{cmp}},
\label{eqn:icp_dynamics_solution}
\]
assuming $\V{r}_{\text{cmp}}$ is held constant throughout $t$.
Using this equation, we can calculate a desired ICP trajectory for walking, given a set of desired footsteps and desired CMP locations in those footsteps.
To more accurately represent human-like walking, we use two CMPs per foot, one in the heel ($\V{r}_{\text{cmp},H}$) and one in the toe ($\V{r}_{\text{cmp},T}$), as shown in \autoref{fig:recursive_icp_plan} by the green circles.
This results in the reference CMP trajectory moving from the heel to the toe in the foot while stepping.

To determine the desired ICP trajectory, we can recurse backward from the final objective location.
This can be done by using the solution to the ICP dynamics in \autoref{eqn:icp_dynamics_solution}, and assuming a static CMP location.
We can define the time spent on the toe-CMP as a fraction of the full step duration, $T_{TH} = \alpha_{TH} T$ and the corresponding time spent on the heel-CMP as $T_{HT} = \left( 1 - \alpha_{TH} \right) T$.
Using this, we can calculate the ICP \say{corner points}, $\V{\xi}_{TH,i}$ and $\V{\xi}_{HT,i}$.
This results in the dark blue trajectories in \autoref{fig:recursive_icp_plan}(a).

To achieve this reference trajectory, however, an instantaneous shift is required from the reference CMP locations, $\V{r}_{\text{cmp},H,i}$ and $\V{r}_{\text{cmp},T,i}$.
Instead, we can smooth these trajectories using third order polynomial interpolation, which guarantees smoothness of the CMP trajectory~\citep{Englsberger_2014}.
The general goal during the transfer state is to shift the desired CMP from the previous toe to the upcoming heel.
As such, we can define the initial ICP location at the start of double support, $\V{\xi}_{\text{iniDS},i}$, and the ICP location at the end of double support, $\V{\xi}_{\text{eoDS},i}$, with respect to the corner point $\V{\xi}_{HT,i}$ as
\[
\begin{aligned}
\V{\xi}_{\text{iniDS},i} &= \V{r}_{\text{cmp},T,i-1} + e^{-\omega_0 T_{\text{iniDS}}} \left( \V{\xi}_{HT,i} - \V{r}_{\text{cmp},T,i-1} \right),
\\
\V{\xi}_{\text{eoDS},i} &= \V{r}_{\text{cmp},H,i} + e^{\omega_0 T_{\text{eoDS}}} \left( \V{\xi}_{HT,i} - \V{r}_{\text{cmp},H,i} \right).
\end{aligned}
\]
The durations to compute these boundary conditions are defined by
$T_{\text{iniDS}} = \alpha_{\text{iniDS}} T_{DS}$ and $T_{\text{eoDS}} = \left( 1 - \alpha_{\text{iniDS}} \right) T_{DS}$, where $T_{DS}$ is the transfer duration.
These knots are shown as the dark red circles in \autoref{fig:recursive_icp_plan}(b).
This spline can then be used to compute the ICP position and velocity as a linear function of the boundary conditions,
\[
\begin{aligned}
\V{\xi}(t) = \V{C}_\xi (t) \V{\Xi}_{\text{bnd}}, & \ \  \dot{\V{\xi}}(t) = \V{C}_{\dot{\xi}}(t) \V{\Xi}_{\text{bnd}}.
\label{eqn:reference_icp_interpolation}
\end{aligned}
\]
Here, the matrices $\V{C}_{\xi}$ and $\V{C}_{\dot{\xi}}$ encode the polynomial values at time $t$.
This results in the light blue and orange colored lines in \autoref{fig:recursive_icp_plan}(b).
The reference CMP trajectory can then be calculated using \autoref{eqn:cmp_definition}.

\begin{figure}[!t]
\centering
\includegraphics[width=3.4in]{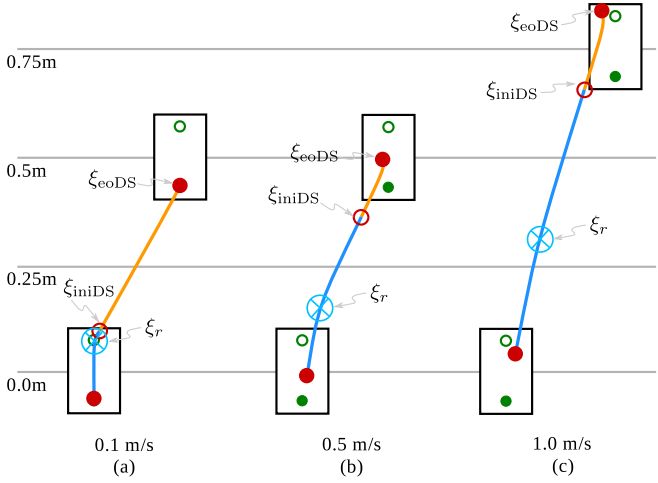}
\vspace{-7mm}
\caption{Diagram showing step plans at different walking speeds. 
The light blue lines represent the ICP trajectory during swing, while the orange lines are during transfer.
}
\vspace{-7mm}
\label{fig:multiple_plans}
\end{figure}

This approach for ICP planning leads to the trajectories shown in \autoref{fig:multiple_plans}, which uses $\alpha_{TH}=\alpha_{\text{iniDS}}=0.5$.
As the walking speed is increased, the resulting plans become more dynamic.
The blue cross represents the desired ICP location half-way through the swing state.
As shown, as the walking speed increases, this ICP location gets further outside of the foot, representing more dynamic walking trajectories.

\subsection{Control}
\label{control}

In our momentum-based control framework, the desired CMP position $\V{r}_{\text{cmp},d}$ is transformed to the desired rate of change of the horizontal linear momentum of the robot by
\[
\dot{\V{l}} = m \omega_0^2 \left( \V{x} - \V{r}_{\text{cmp},d} \right).
\]
This becomes the momentum objective to the whole-body controller described in~\citep{Koolen_2016b}.
$\V{r}_{\text{cmp},d}$ can be calculated using a simple proportional feedback law~\citep{Wiedebach_2016},
\[
\V{r}_{\text{cmp},d} = \V{\xi} - \frac{1}{\omega_0} \dot{\V{\xi}}_r + \V{k}_p \left( \V{\xi} - \V{\xi}_r \right),
\label{eqn:general_feedback_controller}
\]
where $\V{\xi}$ is the measured ICP location.
Inserting the ICP dynamics from \autoref{eqn:icp_dynamics} into \autoref{eqn:general_feedback_controller} yields
\[
\V{r}_{\text{cmp},d} =  \V{k}_\xi  \left( \V{\xi} - \V{\xi}_r \right) + \V{r}_{\text{cmp},r},
\label{eqn:arranged_feedback_controller}
\]
where $\V{k}_\xi = \V{k}_p + \V{1}$, showing that the controller simply adjusts the CMP proportional to the current ICP error.

\section{Swing Speed Up}
\label{swingspeedup}

While in an ideal scenario, humanoid robots do not experience any tracking errors when walking, this is, unfortunately, almost never the case.
Any combination of circumstances can combine to induce these errors, from joint stiction to inaccurate dynamic models to external disturbances.
Most commonly, some form of proportional feedback controller, as in \autoref{eqn:general_feedback_controller}, is employed to correct for this tracking error.
This results in applying additional corrective forces to drive the ICP back to the desired path.

An alternative to providing corrective forces during swing is to adjust the timing of the step.
This is a technique commonly utilized by people; when pushed, we will rapidly put our foot down to recover, in addition to or in place of adjusting the step.
If the error occurs along the current ICP trajectory, this then requires no corrective forces at all, instead only setting the foot down.
Additionally, when combined with step adjustment strategies, step timing can be very effective for assisting in rejecting significant ICP tracking errors.
Due to the exponential relationship between the ICP dynamics and the step time, as shown in \autoref{eqn:icp_dynamics_solution}, the required step adjustment to recover from tracking errors increases exponentially as the swing time increases.
This means that the inverse also holds: decreasing the swing time exponentially decreases the required step adjustment.

\begin{figure}[!t]
\centering
\includegraphics[width=1.6in]{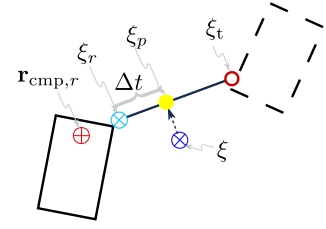}
\vspace{-3mm}
\caption{Illustration of proposed swing speed up calculation.}
\vspace{-5mm}
\label{fig:swing_speed_up}
\end{figure}

We would like to find a time advancement, $\Delta t$, then, such that, at $t^+ = t + \Delta t$, the reference ICP, $\V{\xi}_p$, is as close to the estimated ICP as possible.
From the definition of the ICP dynamics, this value lies on the vector $\V{\xi}_{\text{t}} - \V{\xi}_r$, where $\V{\xi}_{\text{t}}$ is the final ICP location at touchdown.
This is an accurate description of the ICP dynamics, assuming that the location of $\V{r}_{\text{cmp},r}$ does not change during swing; a valid assumption given appropriate planning.
$\V{\xi}$ can be projected onto this vector to find $\V{\xi}_p$ by
\[
\V{\xi}_p = \V{\xi}_r +\left( \V{\xi} - \V{\xi}_r \right)^T \left( \V{\xi}_{\text{t}} - \V{\xi}_r \right) \frac{ \left( \V{\xi}_{\text{t}} - \V{\xi}_r \right) }{ \left\| \V{\xi}_{\text{t}} - \V{\xi}_r \right\| },
\]
as shown in \autoref{fig:swing_speed_up}.

From $\V{\xi}_p$, we can calculate how much further ahead in time that point is using \autoref{eqn:icp_dynamics_solution}, setting the projected ICP as the end condition,
\[
\V{\xi}_p = e^{\omega_0 \Delta t} \left( \V{\xi}_r - \V{r}_{\text{cmp},r} \right) + \V{r}_{\text{cmp},r}.
\]
From here, $\Delta t$ can be solved for by
\[
\Delta t = \frac{1}{\omega} \log_e \left( \frac{ \V{\xi}_p - \V{r}_{\text{cmp},r} }{ \V{\xi}_r - \V{r}_{\text{cmp},r} } \right).
\]
The ICP plan is then advanced to the new time, $t^+$.
To track the swing foot trajectory, however, instead of advancing the time, we calculate a speed up factor $\sigma$ that will cause the remaining duration to pass more quickly. $\sigma$ can be calculated using $\Delta t$ as
\[
\sigma = \frac{ T_{SS} - t}{T_{SS} - t^+}
\]
where $T_{SS}$ is the desired swing time. This approach prevents discontinuities in the desired position for the swing foot. 

\begin{figure}[!t]
\centering
\includegraphics[width=3.3in]{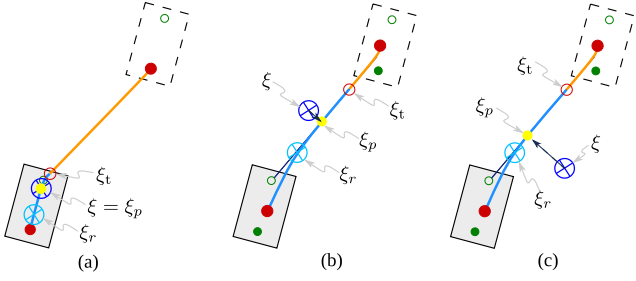}
\vspace{-3mm}
\caption{Speeding up the plan can be very effective when the error is in the direction of the dynamics, as in (a) and (b), but not when it is perpendicular to this motion, as in (c).}
\vspace{-6mm}
\label{fig:speed_up_analysis}
\end{figure}

This control technique is very effective for compensating for errors in the direction of the desired motion, such as being pushed from behind while walking forward, as shown in \autoref{fig:speed_up_analysis}.
If the robot is taking slower steps, as in \autoref{fig:speed_up_analysis}(a), some tracking error purely in the $x$ direction is still on the ICP plan, requiring no corrective forces.
If we take faster steps, as in \autoref{fig:speed_up_analysis}(b), significant forward error still results in relatively small tracking errors once the plan is sped up.
However, this speed up approach is not very effective when the errors are perpendicular to the stepping motion (\autoref{fig:speed_up_analysis}(c)).
Here, the tracking error is only marginally reduced by projecting the ICP onto the plan, requiring either significant corrective forces or step adjustment to compensate.

\section{Step Adjustment}
\label{stepadjustment}

The main objective of the step adjustment algorithm is to combine a proportional feedback controller with one that can adjust the upcoming footsteps.
As we showed in \autoref{swingspeedup}, speeding up the swing is only effective for errors in the directions of the desired ICP dynamics.
When these errors are perpendicular to the dynamics, CMP-based control must be used to try and return the ICP to the reference trajectory.
The control authority granted by moving this value is limited, though, which is equivalent to saying that some tracking errors are too great to return to the nominal plan.
In this case, the only remaining action is to adjust the upcoming footsteps, allowing the footstep to be moved in the direction of the current ICP dynamics.
When combined with the ability speed up the swing plan, this becomes particularly effective, allowing the robot to step quickly to the necessary location to return to the nominal walking plan after $N$ steps.

\subsection{Recursive Dynamics}
\label{recursivedynamics}

Given a footstep plan, we can define $N$ steps to consider for adjustment.
We can then define the first static ICP corner point in the plan, $\V{\xi}_{HT,N+1}$, as $\V{\xi}_{\text{f}}$, from which the local reference value will be defined.
Based on \autoref{eqn:icp_dynamics_solution}, we can see that the ICP corner points are simply linear functions of $\V{\xi}_{\text{f}}$ and the $N$ heel and toe CMP locations.
This is formally defined by
\[
\begin{aligned}
\V{\xi}_{\text{eo}} = & \gamma_{\text{f}} \V{\xi}_{\text{f}} + \textstyle\sum_{i=0}^N \left( \gamma_{T,i} \V{r}_{\text{cmp},T,i}  + \gamma_{H,i} \V{r}_{\text{cmp},H,i} \right), 
\end{aligned}
\label{eqn:fixed_recursive_dynamics}
\]
where $\V{\xi}_{\text{eo}}$ is $\V{\xi}_{TH,0}$ if the robot is currently in the swing state and $\V{\xi}_{HT,0}$ if in transfer.
The scalar multipliers $\gamma_{\text{f}}, \gamma_{T,i},$ and $\gamma_{H,i}$ are computed in Algorithm \autoref{alg:multiplier_calculation}.
If we observe that the CMP locations can be defined relative to footstep positions by
\[
\V{r}_{\text{cmp},T,i} = \V{r}_{\text{off},T,i} + \V{r}_{f,i}, \ \
\V{r}_{\text{cmp},H,i} = \V{r}_{\text{off},H,i} + \V{r}_{f,i}.
\label{eqn:offset_from_foot}
\]
\autoref{eqn:fixed_recursive_dynamics} can then be rewritten as a linear function of the step positions,
\[
\begin{aligned}
\V{\xi}_{\text{eo}} = & \gamma_{\text{f}} \V{\xi}_{\text{f}} + \V{\Xi}_{\text{off}} + \gamma_{T,0} \V{r}_{\text{cmp},T,0} + \gamma_{H,0} \V{r}_{\text{cmp},H,0} \\
& + \textstyle\sum_{i=1}^N \left( \gamma_{T,i} + \gamma_{H,i} \right) \V{r}_{f,i},
\label{eqn:varying_recursive_dynamics}
\end{aligned}
\]
where
\[
\V{\Xi}_{\text{off}} = \textstyle\sum_{i=1}^N \left( \gamma_{T,i} \V{r}_{\text{off},T,i} 
+ \gamma_{H,i} \V{r}_{\text{off},H,i} \right).
\nonumber
\]

\begin{algorithm}
\begin{algorithmic}[1]
\If{Single-Support}
  \State $\V{\xi}_{\text{eo}} = \V{\xi}_{\text{TH},0};$
  \State $\gamma_{T,0} = 1 - e^{-\omega_0 T_{\text{TH},0}};$
  \State $\gamma_{H,0} = 0;$
  \State $\gamma_{\text{f}} = e^{-\omega_0 \left( T_{\text{TH},0} + \sum_{i=1}^N T_i \right) };$
  \For{i = 1,N}
     \State $\gamma_{T,i} = 
        e^{-\omega_0 \left(T_{\text{TH},0} + T_{\text{HT},i} + \sum_{j=1}^{i-1} T_j \right)}
        \left( 1 - e^{-\omega_0 T_{\text{TH},i}}\right);$
     \State $\gamma_{H,i} = 
        e^{-\omega_0 \left(T_{\text{TH},0} +  \sum_{j=1}^{i-1} T_j \right)}
        \left( 1 - e^{-\omega_0 T_{\text{HT},i}}\right);$
  \EndFor
\Else
  \State $\V{\xi}_{\text{eo}} = \V{\xi}_{\text{HT},0};$
  \State $\gamma_{T,0} = e^{-\omega_0 T_{\text{HT},0}}\left( 1 - e^{-\omega_0 T_{\text{TH},0}} \right);$
  \State $\gamma_{H,0} = 1 - e^{-\omega_0 T_{\text{HT},0}};$
  \State $\gamma_{\text{f}} = e^{-\omega_0 \sum_{i=0}^N T_i };$
  \For{i = 1,N}
     \State $\gamma_{T,i} = 
        e^{-\omega_0 \left(T_{\text{HT},i} + \sum_{j=0}^{i-1} T_j \right)}
        \left( 1 - e^{-\omega_0 T_{\text{TH},i}}\right);$
     \State $\gamma_{H,i} = 
        e^{-\omega_0 \sum_{j=0}^{i-1} T_j }
        \left( 1 - e^{-\omega_0 T_{\text{HT},i}}\right);$
  \EndFor
\EndIf
\end{algorithmic}
\caption{Recursive multipliers}
\label{alg:multiplier_calculation}
\end{algorithm}

We can then define the boundary conditions for the splines in transfer and swing for \autoref{eqn:reference_icp_interpolation}.
Using $\V{\xi}_{\text{eo}}$ from \autoref{eqn:varying_recursive_dynamics}, we can define $\V{\Xi}_{\text{bnd}}$ as
\[
\begin{aligned}
\V{\Xi}_{\text{bnd}} = & \V{A}_{\text{F}} \V{\xi}_{\text{eo}} + \V{B}_{T,0} \V{r}_{\text{cmp},T,0} 
+ \V{B}_{H,0} \V{r}_{\text{cmp},H,0},
\label{eqn:spline_boundary_conditions}
\end{aligned}
\]
where $\V{A}_{\text{F}}, \V{B}_{T,0}$, and $\V{B}_{H,0}$ are calculate the boundary conditions from the corner points using the ICP dynamics.

Combining \autoref{eqn:varying_recursive_dynamics} and \autoref{eqn:spline_boundary_conditions} yields $\V{\xi}_r$ as a linear function of the step positions,
\[
\V{\xi}_r = \V{\Phi}_F \V{\xi}_{\text{f}} + \textstyle\sum_{i=1}^N \V{\Gamma}_i \V{r}_{f,i} + \V{\Phi}_{\text{cnst}},
\label{eqn:reference_icp_position}
\]
where
\[
\begin{aligned}
\V{\Phi}_F = & \gamma_{\text{f}} \V{C}_\xi(t^+) \V{A}_{\text{F}},
\\
\V{\Gamma}_i = & \left( \gamma_{T,i} + \gamma_{H,i} \right) \V{C}_\xi(t^+) \V{A}_{\text{F}},
\\
\V{\Phi}_{\text{cnst}} = & \V{C}_\xi(t^+) \left( \V{A}_{\text{F}} \V{\Xi}_{\text{off}} \right.
+ \left( \V{B}_{T,0} + \gamma_{T,0} \V{A}_{\text{F}} \right) \V{r}_{\text{cmp},T,0} \\
& \left. + \left( \V{B}_{H,0} + \gamma_{H,0} \V{A}_{\text{F}} \right) \V{r}_{\text{cmp},H,0} \right).
\nonumber
\end{aligned}
\]

\subsection{Objective Function}
\label{objectivefunction}

\autoref{eqn:arranged_feedback_controller} can be rearranged to yield the corrective CMP action,
\[
\V{\delta} = \V{r}_{\text{cmp},d} - \V{r}_{\text{cmp},r} = \V{k}_\xi\left( \V{\xi} - \V{\xi}_r \right),
\label{eqn:feedback_delta}
\]
where $\V{\delta}$ encodes the amount of corrective forces the robot exerts to try to return to the nominal plan.
By inserting \autoref{eqn:reference_icp_position}, we can see that the feedback action is a function of the current state of the robot $\V{\xi}$, the current time $t^+$, and the upcoming footsteps, $\V{r}_{f,i}$.
Using this, we can define a quadratic program (QP) that optimizes between using feedback control and footstep adjustment, which can be written as
\[
\begin{aligned}
& \min_{\V{r}_{f,i}, \V{\delta} } 
& & \sum_{i=1}^N 
\left\| \V{r}_{f,i} - \V{r}_{f,r,i} \right\|^2_{\V{Q}_{f,i}}
+ \left\| \V{\delta} \right\|^2_{\V{R}} + \left\| \V{\eta} \right\|^2_{\V{Q}_\eta} \\
& \text{subject to}
& & \V{\delta} = \V{k}_\xi \left( \V{\xi} - \V{\Phi}_F \V{\xi}_{\text{f}} - \sum_{i=1}^N \V{\Gamma}_i \V{r}_{f,i} - \V{\Phi}_{\text{cnst}} - \V{\eta} \right),
\end{aligned}
\]
where $\V{Q}_{f,i}$, $\V{Q}_\eta$, and $\V{R}$ are positive definite weighting matrices.
The weight $\V{Q}_{f,i}$ penalizes deviations of the $i^{th}$ footstep position, $\V{r}_{f,i}$, from the $i^{th}$ reference footstep position, $\V{r}_{f,r,i}$.
The weight $\V{R}$ penalizes the use of corrective forces.
$\V{\eta}$ is a slack variable introduced to the dynamics to guard against over constraining the problem, and is minimized by a high weight matrix, $\V{Q}_\eta$.

This controller can be seen to allow the two fundamentally different types of walking to emerge.
If we require that $\V{r}_{f,i} = \V{r}_{f,r,i}$, the robot can no longer adjust its feet, and walks purely by controlling the ICP with the CMP, as with a standard proportional feedback controller.
If $\V{\delta}$ is constrained to equal zero, no correct forces are allowed, and the robot is only allowed to balance through step adjustment, similar to walking with only point feet and a point mass.

In practice, through proper tuning, we can ensure that the robot utilizes its full control authority with the CMP before adjusting the footsteps by setting $\V{Q}_{f,i}$ much greater than $\V{R}$.
The required footstep adjustment has an exponential relation with the tracking error, but only a linear one with $\V{\delta}$.
As such, with proper weighting, increasing $\V{\delta}$ incurs much lower costs than adjusting the footstep.
However, $\V{\delta}$ has limits, which we impose through constraints on the QP in the following section.
This leads to the robot adjusting the footsteps only when absolutely necessary.

\subsection{Problem Constraints}
\label{problemconstraints}

While the CMP is, theoretically, allowed to exit the support polygon through the generation of angular momentum, in practice, this should be used sparingly.
The amount of angular momentum that can be generated is limited, and it must always be \say{paid back} by removing it from the system.
As such, we can constrain the CMP to be within the support polygon by defining a series of equality and inequality constraints
\[
\V{r}_{\text{cmp},d} = \textstyle\sum_c \beta_c \V{r}_c, \ \ 
1 = \textstyle\sum_c \beta_c, \ \
0 \le \beta_c,  \forall c.
\]
This defines the CMP as being a sum of the corner points, $\V{r}_c$, of the polygon.

Additional constraints can be placed on the footstep locations, as long as they represent an affine function
\[
\V{A}_{r,i} \V{r}_{f,i} \le \V{b}_{r,i}, \forall i.
\]
In this work, we used this to define a simple rectangular reachability constraint for the robot.
This formulation can also be used to constrain the footstep location to permissible convex regions, such as the planar regions used in the original footstep planning algorithm.

\section{Results and Discussion}
\label{resultsanddiscussion}

\begin{figure}[!t]
\centering
\includegraphics[width=3.4in]{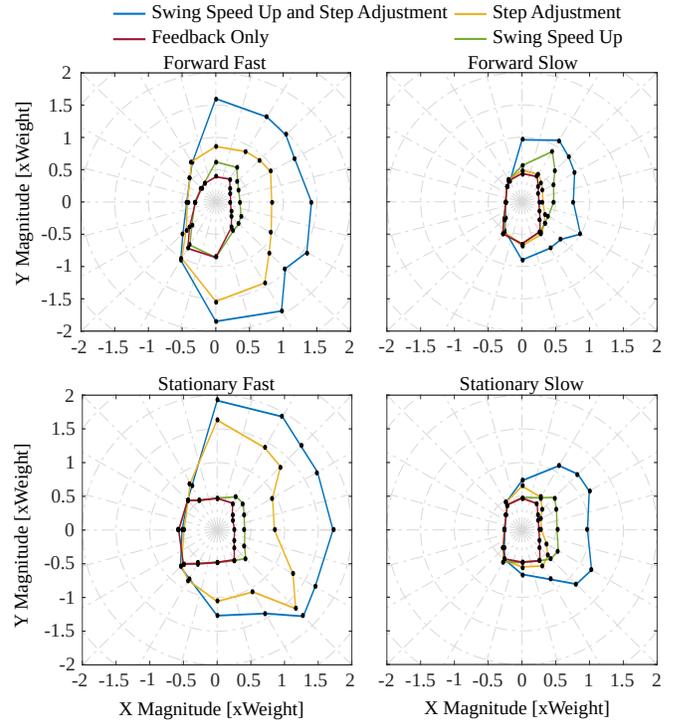}
\vspace{-7mm}
\caption{
Maximum push the robot can recover from and continue walking, at different push angles and step speeds, as a function of the robot weight, using different push recovery methods.
The push is applied to the center of mass for $0.1s$.
Forward steps are $0.5m$ long.
}
\vspace{-6mm}
\label{fig:recoverable_pushes}
\end{figure}

We used the above walking controller both in simulation and on the hardware platform for the Atlas robot.
Using a quad-core 2.7 GHz 3rd generation i7 processor, the QP was solved using a custom active-set solver at in an average $80\mu s$, while the entire algorithm took an average $220 \mu s$, allowing it to be easily solved in real-time.

To explore the effectiveness of different ICP control mechanisms, we conducted simulations comparing the maximum external disturbance that can be recovered from by the four different control mechanisms: proportional feedback only, feedback with step adjustment, feedback with swing speed up, and feedback with step adjustment and swing speed up.
The results of applying disturbances in different directions to different step motions are shown in \autoref{fig:recoverable_pushes}.
Each disturbance was applied to the center of mass of the robot for $0.1s$ halfway through the step.
The step motions included $0.5m$ forward and stationary steps, both fast ($0.95s$) and slow ($2.0s$).
This minimum swing time allowed after speed up was $0.6s$.
The inclusion of additional stabilizing mechanisms (step adjustment, etc.) to the feedback controller was found to improve disturbance rejection, while adding both speed up and step adjustment was consistently the most robust method.
Speed up was generally more effective than step adjustment when walking slowly, as the corresponding required step adjustment was quite large due to the slower step speed.
Exceptions to this are when tracking errors are perpendicular to the dynamics, such as being pushed forward when stepping in place.
As expected, the effectiveness of step adjustment for stabilization was dramatically increased by increasing the step speed.
It is worth noting that the magnitude of recoverable disturbances using only feedback did not significantly change between the different step speeds.
Using both speed up and step adjustment, the largest recoverable disturbance in simulation was 1.92 times its weight, or $2937N$, when stepping quickly in place.

\begin{figure}[!t]
\centering
\includegraphics[width=3.4in]{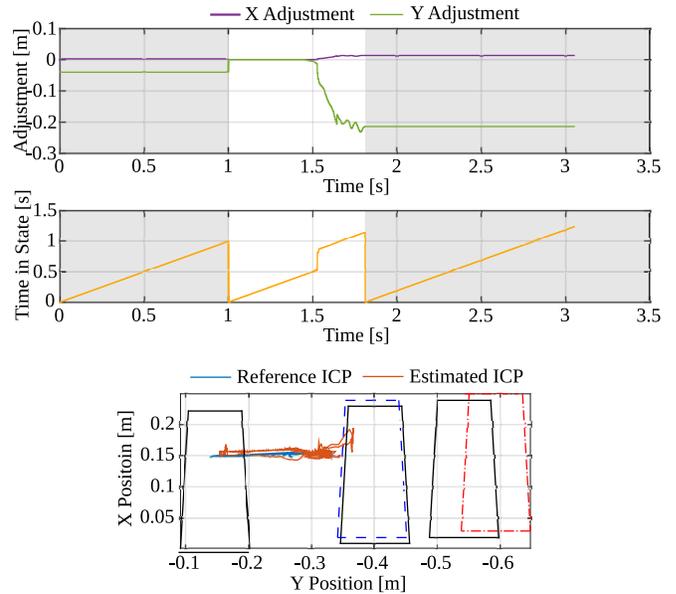}
\vspace{-7mm}
\caption{
Results of applying an outward push when stepping in place. The gray background represents the transfer phase.
The dashed blue foot is the reference footstep, the dashed red footstep is the reference footstep with adjustment, and the black footstep is the actual foot location.
}
\vspace{-4mm}
\label{fig:side_push}
\end{figure}

\begin{figure}[!t]
\centering
\includegraphics[width=3.4in]{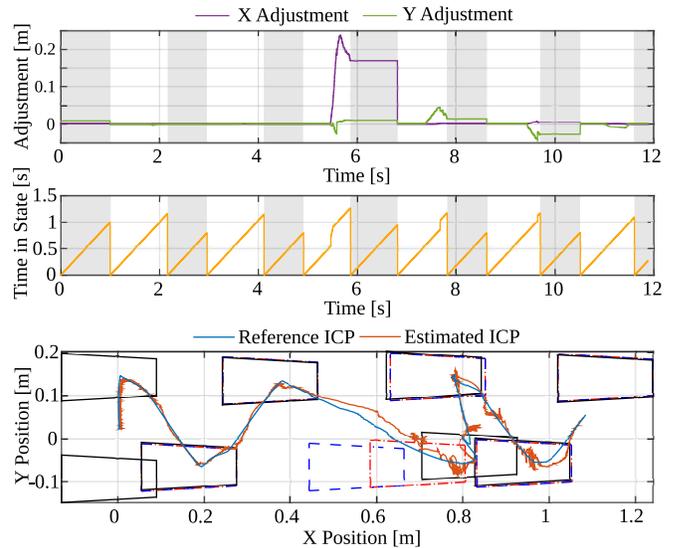}
\vspace{-7mm}
\caption{
Results of applying an forward push when walking forward. 
The gray background represents the transfer phase.
The dashed blue foot is the reference footstep, the dashed red footstep is the reference footstep with adjustment, and the black footstep is the actual foot location.
}
\vspace{-6mm}
\label{fig:forward_push}
\end{figure}

The real robot was also able to successfully use this algorithm to adjust the step timing and locations to compensate for large tracking errors.
We forced these tracking errors by pushing the robot while stepping.
In both presented experiments, the steps durations were $2s$, with $1s$ spent in transfer and $1s$ in swing.
\autoref{fig:side_push} shows the results of applying an outward push when stepping in place.
As can be seen, the reference time is advanced during swing to speed up the ICP trajectory, and the foot is adjusted outward to help maintain balance, with some tracking errors due to the high speed required in the adjustment.
\autoref{fig:forward_push} shows the results of a forward push while the robot is walking.
Again, the swing state is sped up, and the step adjusted in the direction of the push.
Low frequency oscillations in the ICP position occurred after heel strike due to the high speed at which the robot set the foot down, but were quickly damped out.
The impact speed also resulted in additional ICP tracking errors in the direction of the stance foot, but this was easily corrected given the additional control authority during transfer.

While the presented algorithm requires fairly accurate control of the CoP and CMP,
the ability to adjust the step outward based on the ICP dynamics somewhat relaxes this requirement.
By expanding the support polygon, the robot's CoP control authority is less likely to become saturated by operating further from the support polygon boundary, where accuracy is lowest as well.
On the Atlas robot, the CoP is controllable with an accuracy of approximately $2cm$ due to good force control in the ankle joints.
Based on the constraints we have set on the CMP location, this is roughly equivalent to the CMP accuracy.
However, as we are not directly measuring the CMP, there may be unquantified tracking errors caused by unmeasured deviations of the actual CMP from the actual CoP.
Greater control authority could be gained with angular momentum by allowing the CMP to leave the support polygon, as well.
This could be done by adding an additional control variable to the optimization describing deviations from the CMP and the CoP, and then minimizing this deviation while constraining the CoP.

The proposed algorithm does not significant provide improvements against tracking errors in the inward direction of the step.
The step reachability polygon does not allow for any crossover of the steps, simply constraining them to a minimum inward position.
This is due to the difficulties in defining a reachability region that enables crossover while maintaining convexity, as well as range of motion limitations.
By defining multiple possible reachability constraints and selecting the active one based on the current step type and tracking errors, however, crossover could be possible.

A variety of factors led to performance limitations of this controller when ported from simulation to hardware.
These include:
Errors in the robot model.
When using an inverse dynamics-based approach, model accuracy greatly affects the resulting ground reaction forces.
If the controller cannot effectively achieve the CoP at the support polygon edge, it will not be able to as successfully mitigate tracking errors;
Actuator speed and torque limits, which bounds how quickly the robot can step.
By increasing this step speed, we expect the effectiveness of step adjustment algorithms to greatly improve, as illustrated in \autoref{fig:recoverable_pushes};
Sensor noise, which greatly affects the precision of the ICP calculation.
Measurement uncertainty further exacerbates inaccuracies in the inverse-dynamics calculation, as well as other task-space controllers.

\section{Conclusion}
\label{conclusion}

The ability to robustly recover from large tracking errors is essential to improving the capabilities of humanoid robots, and represents a critical step forward in enabling them to competently function in uncertain environments.
In this work, we presented a new approach for adjusting both step timing and locations to reject external disturbances and their corresponding tracking errors.
By including step timing adjustment, the required step adjustment to reject errors is exponentially decreased.
Our algorithm formulates this problem in a highly efficient manner, allowing it to be solved quickly in real-time.
In the future, we hope to incorporate angular momentum in the algorithm to further increase the control authority available to the robot.
We also plan to integrate the step timing adjustment into the optimization algorithm, borrowing from the gradient descent approaches used by air vehicles~\citep{Mellinger_2011}.
We will additionally include environmental information to allow the step adjustment algorithm to be used effectively in dynamic and cluttered environments.


\ifx\myappendix\undefined
\else

\section*{APPENDIX}
\input{\myappendix}

\fi

\ifx\myacknowledgments\undefined
\else

\section*{ACKNOWLEDGMENT}
\input{\myacknowledgments}

\fi

\ifx\bibliocommand\undefined
\else
\printbibliography
\fi

\end{document}